\documentclass[unnumsec,webpdf,contemporary,large]{oup-authoring-template}%

\graphicspath{{Fig/}}

\theoremstyle{thmstyleone}%
\theoremstyle{thmstyletwo}%
\theoremstyle{thmstylethree}%

\usepackage{xcolor}         
\usepackage{transparent}    
\usepackage{booktabs}       
\usepackage{multirow}       
\usepackage{tabularx}       
\usepackage{adjustbox}      
\usepackage{caption}        
\usepackage{makecell}       
\usepackage{threeparttable} 
\usepackage{float}          

\begin{document}

\journaltitle{Journal Title Here}
\DOI{DOI HERE}
\copyrightyear{2022}
\pubyear{2019}
\access{Advance Access Publication Date: Day Month Year}
\appnotes{Paper}

\firstpage{1}


\title[Short Article Title]{OEMA: Ontology-Enhanced Multi-Agent Collaboration Framework for Zero-Shot Clinical Named Entity Recognition}

\author[1]{Xinli Tao, B.S.}
\author[1,$\ast$]{Xin Dong, M.S.}
\author[1,$\ast$]{Xuezhong Zhou, Ph.D.}

\authormark{Tao et al.}

\address[1]{\orgdiv{Department of Artificial Intelligence, Beijing Key Lab of Traffic Data Analysis and Mining}, 
\orgname{School of Computer Science \& Technology, Beijing Jiaotong University}, 
\orgaddress{\street{Beijing}, \postcode{100044}, \country{China}}}

\corresp[$\ast$]{Corresponding authors. \href{mailto:xzzhou@bjtu.edu.cn}{xzzhou@bjtu.edu.cn}; \href{mailto:x_dong@bjtu.edu.cn}{x\_dong@bjtu.edu.cn}.}








\abstract{
\textbf{Objective:} With the rapid expansion of unstructured clinical texts in electronic health records (EHRs), clinical named entity recognition (NER) has become a crucial technique for extracting medical information. However, traditional supervised models such as CRF and BioClinicalBERT suffer from high annotation costs. Although zero-shot NER based on large language models (LLMs) reduces the dependency on labeled data, challenges remain in aligning example selection with task granularity and effectively integrating prompt design with self-improvement frameworks.\\[3pt]
\textbf{Materials and Methods:} To address these limitations, we propose OEMA, a novel zero-shot clinical NER framework based on multi-agent collaboration. OEMA consists of three core components: (1) a self-annotator that autonomously generates candidate examples; (2) a discriminator that leverages SNOMED CT to filter token-level examples by clinical relevance; and (3) a predictor that incorporates entity-type descriptions to enhance inference accuracy.\\[3pt]
\textbf{Results:} Experimental results on two benchmark datasets, MTSamples and VAERS, demonstrate that OEMA achieves state-of-the-art performance under exact-match evaluation. Moreover, under related-match criteria, OEMA performs comparably to the supervised BioClinicalBERT model while significantly outperforming the traditional CRF method.\\[3pt]
\textbf{Discussion:} OEMA combines ontology-guided reasoning with multi-agent collaboration to address two key challenges in zero-shot clinical NER: granularity mismatch and prompt–self-improvement integration. Its ontology-based filtering reduces noise and enhances semantic alignment, showing strong performance across different LLMs. \\[3pt]
\textbf{Conclusion:} OEMA improves zero-shot clinical NER, achieving near-supervised performance under related-match criteria. Future work will focus on continual learning and open-domain adaptation to expand its applicability in clinical NLP.
}


\maketitle

\section{Introduction}
Electronic health records (EHRs) contain vast amounts of unstructured clinical information, including clinical notes, and this type of information is of great value to clinical experts \cite{b1_kollapally2024using}, for which many studies have been devoted to the problem of clinical information extraction \cite{b2_landolsi2023information}. A key aspect in the extraction of clinical information is named entity recognition (NER), which is focused on identifying specific concepts such as medical problems, treatments, and examinations \cite{b3_nadkarni2011natural}.

Early clinical natural language processing systems typically relied on predefined lexical resources and syntactic/semantic rules derived from manual analysis of large amounts of text \cite{b4_wang2018clinical}. In the last decade, machine learning-based approaches (e.g., CRF \cite{b5_jiang2011study}) have become increasingly popular \cite{b6_huang2015bidirectional}. In recent years, migration learning-based models (e.g., BioClinicalBERT \cite{b7_alsentzer2019publicly}) have been applied to the task of clinical named entity recognition, showing improved performance with fewer annotated samples \cite{b8_bayat2023survey}, but also requiring time-consuming development of annotated corpora by clinical experts \cite{b9_10.1145/3458754}.

Recently, with the advancement of large language models (LLMs), the task of NER has also ushered in new research directions and application possibilities \cite{b10_fi15060192, b11_touvron2023llama, b12_chowdhery2023palm}. 
Owing to their vast search space and large-scale pre-training data, LLMs show great potential for zero-shot NER, as evidenced by recent empirical studies \cite{b13_xie-etal-2023-empirical}.
Existing work has proceeded along two main avenues: first, prompt-engineering methods that construct more effective zero-shot prediction prompts to enhance LLMs’ generalization to out-of-distribution (OOD) knowledge \cite{b14_zhao2024effective, b15_hu2024improving, b16_wei2023chatie}; and second, self-improvement frameworks that use the LLM’s own generated self-consistency scores \cite{b17_wang2022self} to automatically label unlabeled data, thereby building a self-annotated corpus and improving performance at inference time through self-annotated few-shot in-context learning (ICL) \cite{b18_xie-etal-2024-self, b19_wang2024reversener}. As shown in Fig.~\ref{fig:Challenge_description}, despite these advances, current zero-shot NER methods still face two challenging problems:

\begin{figure*}[!t]     
  \raggedleft          
  \includegraphics[width=\linewidth]{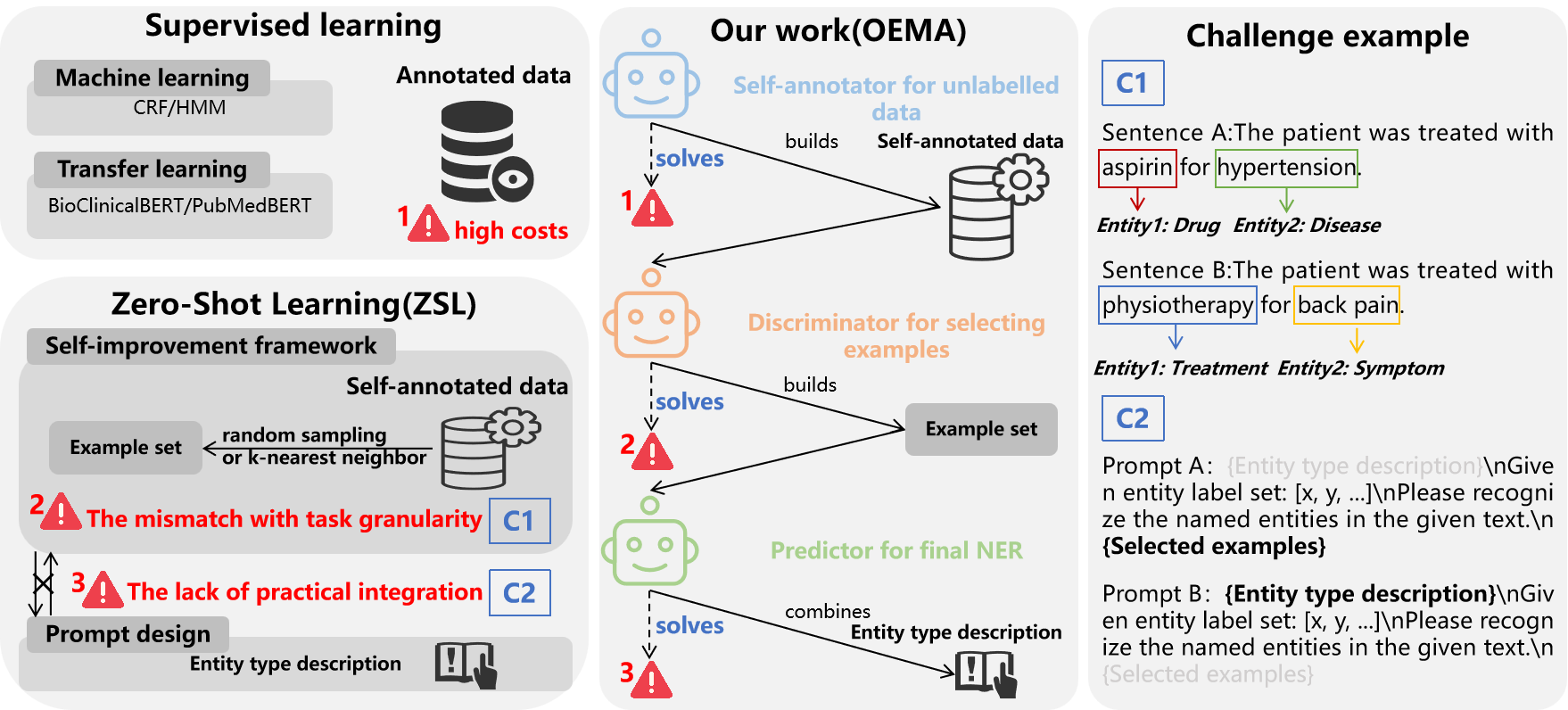}
  \caption{
  Challenge Analysis Diagram. In zero-shot learning, OEMA tackles two key challenges: (1) the mismatch between example selection and task granularity, and (2) the lack of effective integration between prompt design and the self-improvement framework.
  }
  \label{fig:Challenge_description}
\end{figure*}

\begin{itemize}
\item \textbf{Challenge 1: The mismatch between example selection and task granularity}. In self-improvement frameworks, self-annotated examples must be added to prompts to guide the LLM. However, example selection is based on shallow, sentence-level strategies (e.g., random sampling or k-nearest neighbors \cite{b18_xie-etal-2024-self}), whereas NER is a token-level task—relying on sentence-level semantic similarity is therefore inappropriate \cite{b21_wang-etal-2025-gpt}.

\item \textbf{Challenge 2: The lack of practical integration between prompt design and self-improvement frameworks}. Although existing studies have suggested that self-improvement frameworks are “prompt-agnostic” and, in theory, can incorporate any advanced prompt design methods to boost performance \cite{b18_xie-etal-2024-self}, this has not been empirically validated, causing advanced prompts to fall short of realizing their potential within self-annotation frameworks \cite{b22_zhang-etal-2025-survey}.
\end{itemize}

To address the above challenges, we propose OEMA, a zero-shot clinical NER framework illustrated in Fig.~\ref{fig:OEMA_framework}, which comprises three collaborative agents: a self-annotator for unlabeled data, a discriminator for selecting examples, and a predictor for final NER. First, the self-annotator uses LLMs to label an unlabeled corpus, thereby constructing a self-annotated corpus \cite{b18_xie-etal-2024-self}. Next, the discriminator retrieves $K$ candidate examples via cosine-similarity–based search, then applies a dedicated ICL method to extract top-level SNOMED CT concepts and their source text spans from both the test sentence and each example. It assigns each example a usefulness score with respect to the target input and selects the top $k$ examples ($K > k$) as the final example set. Finally, the predictor performs inference on the target test sentence using a prompt that combines entity-type description with the self-annotated few-shot examples (built from that example set). Experimental results on two benchmark datasets demonstrate that OEMA achieves state-of-the-art performance, while ablation experiment, hyperparameter tuning, and case study jointly validate its effectiveness and interpretability.

In summary, the main contributions of this study can be distilled into three aspects:
\begin{enumerate}
\item To address Challenge 1, we propose a medical‐ontology‐driven, token‐level example selection strategy. This approach refines self‐annotated example filtering from the traditional sentence level to a finer-grained token level, and employs a dynamic scoring mechanism to select examples that are highly relevant to the target test sentence, thereby enhancing the guidance provided to the LLM. This strategy effectively narrows the gap between example selection and task granularity, enabling more fine-grained and task-aligned example selection, particularly suitable for clinical NER tasks.
\item To address Challenge 2, we design the OEMA framework under a multi-agent collaborative mechanism, integrating advanced prompt design with the self-improvement framework. The former introduces type-level semantic priors through entity-type descriptions, effectively compensating for the generalization limitations of the latter’s self-annotated few-shot learning. This results in a dual prompting strategy—“type priors + structured examples”—which effectively alleviates the performance bottleneck between prompt design and the self-improvement framework. To the best of our knowledge, this is the first work to explicitly emphasize the synergy between type priors and structured examples within a multi-agent collaborative framework.
\item Benchmark experiments demonstrate that OEMA outperforms mainstream approaches in zero-shot settings. Ablation studies validate the synergistic effect of combining entity-type prompts with self-annotated few-shot prompts, while case analyses further highlight the critical role of high-quality self-annotated examples in driving performance gains. Additionally, the results provide concrete evidence of the fine-grained collaboration achieved by the multi-agent design in orchestrating prompt design and self-improvement within the OEMA framework.
\end{enumerate}

\section{Related work}
\textbf{Clinical named entity recognition.}
Early approaches to clinical NER mainly relied on manually crafted linguistic rules and curated dictionaries, built through labor-intensive text analysis by domain experts \cite{b4_wang2018clinical}. Over time, research in clinical NER has shifted toward data-driven, machine learning–based techniques, which have demonstrated greater adaptability and scalability \cite{b6_huang2015bidirectional}. Well-known clinical information extraction frameworks such as cTAKES and CLAMP now employ hybrid architectures that combine rule-based modules with machine learning components \cite{savova2010mayo}. More recently, transformer-based LLMs have revolutionized clinical natural language processing (NLP). Models such as Bidirectional Encoder Representations from Transformers (BERT) have become the foundation for capturing contextual meaning in unstructured clinical text \cite{devlin2019bert}. Building upon BERT, several domain-specific adaptations—such as BioBERT and PubMedBERT, trained on biomedical research literature, and ClinicalBERT, trained on the MIMIC-III clinical dataset—have been introduced \cite{lee2020biobert, b9_10.1145/3458754, huang2019clinicalbert}. Through transfer learning, these specialized models can be fine-tuned for clinical NER tasks, achieving state-of-the-art performance even with relatively limited annotated data.Despite these advances, a persistent challenge remains—the creation of large, high-quality annotated corpora, which demands significant time and expert involvement \cite{lee2020biobert, b9_10.1145/3458754, huang2019clinicalbert}.

\textbf{Zero-shot named entity recognition.} With the emergence of LLMs such as GPT, zero-shot NER has become a promising alternative to supervised approaches. Owing to their large-scale pretraining and broad world knowledge, LLMs can perform entity recognition tasks through prompt-based instruction without explicit fine-tuning \cite{b13_xie-etal-2023-empirical}. Research in this area mainly follows two directions: the first focuses on prompt engineering, which aims to design effective task descriptions and input templates to guide LLMs toward accurate entity extraction. For example, IILLM \cite{b15_hu2024improving} reformulates the NER task as an HTML-style code generation problem, incorporating annotation guidelines and error analysis instructions to improve structure awareness. Similarly, Chatie \cite{b16_wei2023chatie} treats information extraction as an interactive dialogue process, leveraging conversational cues to enhance reasoning and flexibility. Other studies such as Zhao et al. \cite{b14_zhao2024effective} explore optimizing in-context examples to align model behavior with human annotation conventions. The second research direction emphasizes self-improvement or self-annotation mechanisms, where LLMs iteratively generate pseudo-labeled examples to bootstrap their own in-context learning. Xie et al. \cite{b18_xie-etal-2024-self} introduced the Self-Improving LLM (SILLM) framework, which uses self-consistency voting \cite{b17_wang2022self} and example selection to improve NER without external labels, while Wang et al. \cite{b19_wang2024reversener} extended this idea through Reversener, employing self-generated examples for robust zero-shot entity detection.

Compared with traditional clinical NER, zero-shot named entity recognition offers a significant advantage by eliminating the need for costly manual annotation, enabling rapid adaptation to new entity types and datasets. Despite these advances, existing zero-shot NER approaches still face two critical challenges. First, there remains a persistent mismatch between example selection granularity and task level \cite{b21_wang-etal-2025-gpt}. Most self-improvement frameworks rely on sentence-level similarity retrieval, lacking fine-grained modeling of token-level semantic relevance. In clinical NER, such coarse selection often introduces noisy examples, impairing entity boundary precision and semantic consistency. Second, there is limited integration between prompt engineering and self-improvement frameworks \cite{b22_zhang-etal-2025-survey}. Existing studies typically treat these components independently—prompt engineering focuses on designing task-specific prompts, while self-improvement emphasizes optimization through self-generated data. Without explicit synergy, these methods struggle to achieve both generalization and semantic constraint in unsupervised settings. To address these issues, a collaborative system integrating token-level example selection with type-aware prompting under a tri-agent architecture consisting of a self-annotator, discriminator, and predictor is proposed.

\section{Methods}
\subsection{Problem definition}
We now formally define the zero-shot NER task. Given an input sentence $x = (w_1, w_2, \cdots, w_n)$, where $w_i$ represents the $i$-th token in the sentence, the goal of zero-shot NER is to identify all semantically meaningful named entities in the sentence $x$ \textit{without using any manually annotated training examples}, and to structure the identified entities in the following output form:
\begin{equation}\label{eq:zsner}\tag{1}
y \;=\;\{(e,t)\mid e \subseteq x,\ t \in \mathcal{T}\},
\end{equation}
Here, $e$ denotes an entity span, i.e., a contiguous segment of text in sentence $x$; $t$ is the type of the entity, which belongs to a predefined set of entity types $\mathcal{T}$. The output structure $y$ consists of multiple pairs $(e, t)$, representing all identified entities and their corresponding types.

\subsection{OEMA framework}
\begin{figure*}[htbp]
    \centering
    \includegraphics[width=.96\linewidth]{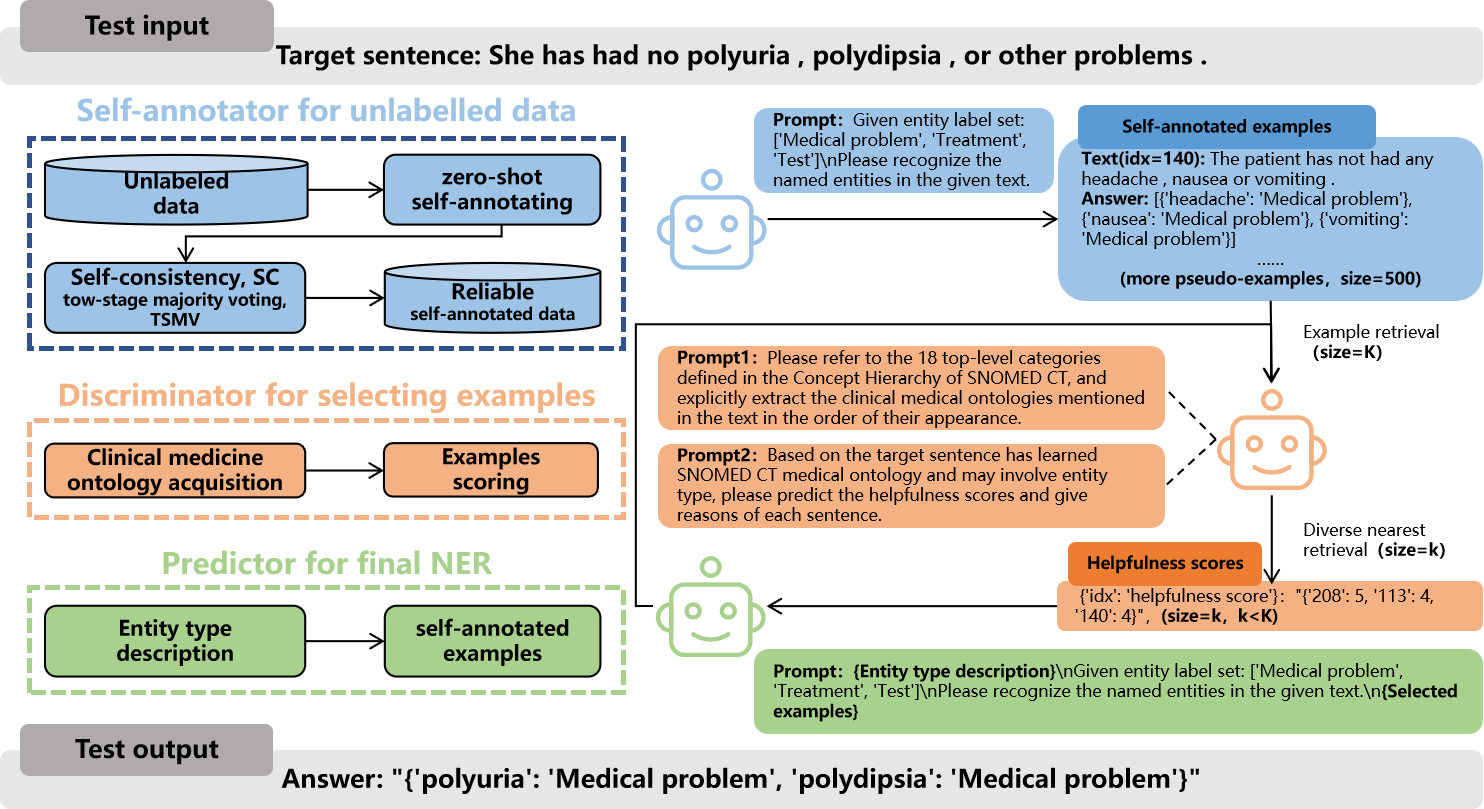}
    \caption{Framework of OEMA: (1) Self-Annotator creates a corpus from unlabeled data; (2) Discriminator scores token-level examples via top-level SNOMED CT ontologies; (3) Predictor fuses entity-type descriptors with selected examples to yield the NER output. Agents are grouped by function (dashed boxes) and connected by arrows to show the execution order.}
    \label{fig:OEMA_framework}
\end{figure*}

As shown in Fig.~\ref{fig:OEMA_framework}, OEMA comprises three core agents: the self-annotator, which constructs a self-annotated corpus from unlabeled data; the discriminator, which scores and ranks examples using the top-level SNOMED CT ontology; and the predictor, which combines self-annotated few-shot prompts with entity-type descriptions to produce the final NER output.

\subsubsection{Self-annotator for unlabelled data}
In the zero-shot NER setting, we only have access to an unlabeled corpus. Therefore, we introduce a self-annotator inspired by the self-improvement strategy \cite{b18_xie-etal-2024-self}, which guides the reasoning process of LLMs. Specifically, the self-annotator uses zero-shot prompt to annotate the unlabeled corpus, thereby constructing a self-annotated corpus.
For each unlabeled sample $x_i$, we use zero-shot prompt to generate predictions based on the LLMs.  
This process is defined by (\ref{eq:self-annotate}) as follows,
\begin{equation}\label{eq:self-annotate}\tag{2}
y_i = \arg\max_{\mathbf{y}} P_s(\mathbf{y} \mid \mathcal{T}_s, x_i)
\end{equation}
where $\mathcal{T}_s$ is the prompt template used for self-annotation. 
For an example of $\mathcal{T}_s$, refer to Table.~A1 in the Supplementary Material. 
$P_s$ denotes the output probability from the self-annotator. 
The prediction result $y_i$ is structured as shown in (\ref{eq:yi_format}), which consists of a set of entity-type pairs. Here, $l$ denotes the number of predicted entities.


\begin{equation}\label{eq:yi_format}\tag{3}
y_i = \left\{ (e_j^i, t_j^i) \right\}_{j=1}^{l}
\end{equation}

To improve the reliability of annotations, we employ self-consistency \cite{b17_wang2022self} and adopt a two-stage majority voting strategy \cite{b13_xie-etal-2023-empirical}. We sample multiple responses from LLMs. In the first stage, if a candidate mention appears in more than half of all responses, it is considered an entity; otherwise, it is discarded. In the second stage, for each mention retained from the first stage, we determine its entity type based on the majority opinion among the retained mentions, and this is taken as the final predicted entity type.

The self-annotator provides the self-improvement framework with essential, high-quality foundational data, significantly reducing dependence on manual annotation and its associated costs.

\subsubsection{Discriminator for selecting examples}
Through the aforementioned self-annotator agent, we obtain a self-annotated corpus, and to further identify high-quality self-annotated examples, we design a discriminator for selecting examples.
When the target sentence $x^q$ arrives, we adopt a diversified nearest‐neighbor approach. First, we retrieve $K$ relevant examples $S_d = \{(x_i, y_i)\}_{i=1}^K$ from the self‐annotated corpus based on cosine similarity, and then select the top $k$ examples with the highest usefulness scores.
To enable the model to select examples based on token‐level similarity grounded in medical ontologies, we design a specialized ICL for clinical ontology extraction. Let the self‐annotated example set $S_d$ and the target sentence $x^q$ form the input set $I_u \in S_d \cup \{x^q\}$. For each sample $\mathfrak{x}_i \in I_u$, we use few‐shot example prompt to generate results. The formula for this procedure is as follows,

\begin{equation}\label{eq:ontology_extraction}\tag{4}
o_i \;=\;\arg\max_o \;P_e\bigl(y \mid T_e,\,\mathfrak{x}_i\bigr)
\end{equation}
where $T_e$ denotes the ICL prompt template for clinical ontology extraction. 
For an example of $T_e$, refer to Table.~A2 in the Supplementary Material. 
$P_e(\cdot)$ represents the output probability of the clinical ontology extraction.


When examples retrieved by a shallow similarity–based strategy may be highly irrelevant to the target sentence and severely mislead the LLMs’ predictions \cite{b18_xie-etal-2024-self}, we address this issue by assigning a helpfulness score to each example to automatically assess its contribution to the prediction of the target test sentence. Specifically, given the example set $S_d$ for the target sentence $x^q$ and the generated clinical ontology set $O_d = \{(\mathfrak{x}_i, o_i)\}_{i=1}^{K+1}$, the corresponding helpfulness scores $\{h_i\}_{i=1}^K$ are predicted by the following equation,

\begin{equation}
\label{eq:example_scoring}\tag{5}
h_i = \arg\max_{h} \;P_d\bigl(y \mid T_d,\,S_d,\,O_d,\,\mathfrak{x}_i,\,x^q\bigr)
\end{equation}
where $T_d$ denotes the ICL prompt template for clinical ontology extraction. 
For an example of $T_d$, refer to Table.~A3 in the Supplementary Material. 
$P_d(\cdot)$ represents the probability used for scoring each example.

Finally, we sort the helpfulness scores $\{h_i\}_{i=1}^K$, and select the top $k$ examples with the highest scores as the final example set $S_o = \{(x_i, y_i)\}_{i=1}^k$, thus completing the entire diversified nearest neighbor selection process.

This token-level, ontology-enhanced selection mechanism tackles challenge 1 by retrieving fine-grained, semantically relevant examples, offering the LLM precise guidance and avoiding misalignment from coarse, sentence-level matching.

\subsubsection{Predictor for final NER}
Building on the previous two agents (self-annotator and discriminator), the key is now to fully exploit the dual prompting strategy of “type prior and structured examples”. Therefore, we introduce a predictor for final NER.
The predictor performs few‐shot ICL by combining the $k$ examples $S_o$ with the target sentence $x^q$. The final prediction is given as follows,

\begin{equation}\label{eq:final_prediction}\tag{6}
y^q = \arg\max_y P_o\bigl(y \mid T_o,\,S_o,\,x^q\bigr)
\end{equation}
where $T_o$ denotes the ICL prompt template that incorporates entity‐type descriptions. 
For an example of $T_o$, refer to Table.~A4 in the Supplementary Material. 
$P_o(\cdot)$ represents the predictor’s output probability.


This fusion of entity-type descriptions and self-annotated examples addresses Challenge 2 via a multi-agent workflow: the self-annotator generates candidate knowledge from unlabeled data, the discriminator filters high-quality examples, and the predictor performs final NER. This design allows the LLM to integrate type priors with structured examples, achieving state-of-the-art performance in clinical NER.

\section{EXPERIMENTS AND RESULTS}

To evaluate the effectiveness of the proposed OEMA, our goal is to verify it by answering the following research questions.

\textbf{RQ1:} Does OEMA’s exact‐match performance on zero-shot clinical named entity recognition tasks outperform the state-of-the-art methods? (See Section~\ref{sec:3.4.1})

\textbf{RQ2:} How does OEMA’s performance on zero-shot clinical named entity recognition tasks compare to that of traditional supervised learning? (See Section~\ref{sec:3.4.1})

\textbf{RQ3:} What are the specific contributions of OEMA’s self-annotated few-shot prompting strategy versus its entity-type description prompting strategy? (See Section~\ref{sec:3.4.2})

\textbf{RQ4:} How are the two key hyperparameters, $K$ and $k$, tuned in OEMA’s diversified nearest-neighbor method? (See Section~\ref{sec:3.4.3})

\textbf{RQ5:} What is the detailed process by which OEMA improves the evaluation metrics? (See Section~\ref{sec:3.4.4})

\subsection{Dataset}
This study utilizes two clinical NER datasets:
(i) MTSamples\cite{b23_uzuner20112010}, which contains 163 synthetic discharge summaries annotated according to the 2010 i2b2 guidelines, used for extracting medical problems, treatments, and tests;
(ii) VAERS\cite{b24_du2021extracting}, which includes 91 publicly available VAERS safety reports, used for identifying neurological disorder events.

The datasets were split into training, validation, and test sets. The self‐annotated samples for the OEMA framework were drawn from the first 500 instances of the training set; the validation set was aligned with those used for the CRF and BioClinicalBERT models; and the test set was used for performance evaluation and comparison. Table.~\ref{tab:datasets_entities} presents the entity statistics for each dataset.

\begin{table}[htbp]
    \centering
    \caption{Datasets and Entities Distribution.}
    \label{tab:datasets_entities}
    \adjustbox{width=\columnwidth}{
    \begin{tabular}{l l r r r r}
        \toprule
        \textbf{Datasets} & \textbf{Entities} & \textbf{Train} & \textbf{Valid} & \textbf{Test} & \textbf{Total} \\
        \midrule
        MTSamples\cite{b23_uzuner20112010} & Medical problem & 538 & 203 & 199 & 940 \\
                 & Treatment & 149 & 43 & 35 & 227 \\
                 & Test & 120 & 39 & 50 & 209 \\
        \midrule
        VAERS\cite{b24_du2021extracting} & Investigation & 148 & 29 & 59 & 236 \\
              & Nervous adverse event & 406 & 83 & 162 & 651 \\
              & Other adverse event & 301 & 62 & 167 & 530 \\
              & Procedure & 338 & 57 & 126 & 521 \\
        \bottomrule
    \end{tabular}
    }
    \begin{tablenotes}
        \item Abbreviation: VAERS—vaccine adverse event reporting system.
    \end{tablenotes}
\end{table}

\subsection{Baseline}
To compare model performance, we adopted the following baseline methods:
(i) Vanilla \cite{b13_xie-etal-2023-empirical} employs a straightforward and commonly used prompting strategy that directly asks the large language model to extract entity labels from the input text.
(ii) IILLM \cite{b15_hu2024improving} transforms the zero-shot NER task into an HTML code generation task using a carefully designed prompting strategy that includes task descriptions, annotation guidelines, and error analysis instructions.
(iii) SILLM  \cite{b18_xie-etal-2024-self} applies a self-improvement framework that leverages a self-annotated corpus to stimulate the LLM’s self-learning capabilities in zero-shot NER.

\subsection{Evaluation criteria and experimental setup}
Model performance was evaluated using the 2010 i2b2 challenge evaluation script \cite{b23_uzuner20112010}, computing precision (P), recall (R), and F1 score (F1) under both exact-match (exact boundary and entity type agreement) and relaxed-match (same entity type with text overlap). Following Xie et al. \cite{b18_xie-etal-2024-self}, the self‐consistency score was computed with temperature = 0.7 over 5 sampled outputs, and diversified K‐nearest neighbors (K = 12) were used to generate k = 3 self‐annotated examples.

For a fair comparison, all baselines were re-implemented using gpt-3.5-turbo-0125 (to address version discrepancies in the original papers), and both OEMA and all baselines were also implemented on gemini-2.5-flash-preview-04-17 to evaluate inference capabilities. Text embeddings were generated using text-embedding-ada-002. In the experiments, the self-annotator and predictor varied based on the experimental setup (gpt-3.5 or gemini), while the discriminator was consistently fixed to gpt-3.5-turbo-0125 to balance performance and cost.

\subsection{Results}\label{sec:results}
\subsubsection{Main Results} \label{sec:3.4.1}

\begin{table*}[htbp]
\centering
\caption{MT Samples and VAERS Exact-Match Results. Numbers in \textbf{bold} are the highest results for the corresponding dataset, while numbers \underline{underlined} represent the second-best results. Significant improvements against the best-performing baseline for each dataset are marked with $^\ast$ (t-test, $p < 0.05$).}
\label{tab:main_results_1}
\begin{tabular}{l l c c c c c c}
\toprule
 & & \multicolumn{3}{c}{\textbf{MT Samples}} & \multicolumn{3}{c}{\textbf{VAERS}} \\
\cmidrule(lr){3-5} \cmidrule(lr){6-8}
\textbf{Backbones} & \textbf{Models} & \textbf{P} & \textbf{R} & \textbf{F1} & \textbf{P} & \textbf{R} & \textbf{F1} \\
\midrule
\multirow{4}{*}{gpt-3.5-turbo} 
 & Vanilla & 41.5 & 33.6 & 37.1 & 36.4 & 29.6 & 32.7 \\
 & SILLM & \underline{48.0} & \underline{43.1} & \underline{45.4} & 39.1 & \underline{36.7} & \underline{37.9} \\
 & IILLM & 46.7 & 40.1 & 43.2 & \underline{44.7} & 23.0 & 30.3 \\
 & OEMA (ours) & \textbf{49.4}\,$^\ast$ & \textbf{54.1}\,$^\ast$ & \textbf{51.6}\,$^\ast$ & \textbf{46.4}\,$^\ast$ & \textbf{49.6}\,$^\ast$ & \textbf{48.0}\,$^\ast$ \\
\midrule
\multirow{4}{*}{gemini-2.5-flash}
 & Vanilla & 54.3 & 62.2 & 58.0 & 48.0 & 49.2 & 48.6 \\
 & SILLM & 57.7 & 63.3 & 60.4 & 48.1 & 51.6 & 49.8 \\
 & IILLM & \underline{60.1} & \underline{66.9} & \underline{63.3} & \underline{49.5} & \underline{58.4} & \underline{53.6} \\
 & OEMA (ours) & \textbf{61.5}\,$^\ast$ & \textbf{68.2}\,$^\ast$ & \textbf{64.7}\,$^\ast$ & \textbf{53.3}\,$^\ast$ & \textbf{61.0}\,$^\ast$ & \textbf{56.9}\,$^\ast$ \\
\midrule
\multirow{2}{*}{supervised learning}
 & CRF & 51.1 & 68.1 & 58.4 & 47.3 & 59.1 & 52.5 \\
 & BioClinicalBERT & 78.5 & 78.5 & 78.5 & 69.8 & 64.0 & 66.8 \\
\bottomrule
\end{tabular}
\end{table*}

We first focus on RQ1, which compares the performance of the OEMA model against existing baselines on clinical text information extraction. Table.~\ref{tab:main_results_1} shows that under the exact-match F1 metric, OEMA significantly outperforms all baseline models on both datasets, demonstrating its superior clinical entity recognition capability. When using gpt-3.5 as the backbone LLM, OEMA achieves an exact-match F1 score that is 6.2\% higher than the best baseline on the MTSamples dataset and 10.1\% higher on the VAERS dataset. With gemini-2.5-flash as the backbone LLM, OEMA likewise maintains its lead, improving F1 scores by 1.4\% on MTSamples and 3.3\% on VAERS compared to the top baseline. These figures clearly demonstrate the effectiveness of the OEMA framework.

Further investigation revealed that the choice of LLM backbone has a significant impact on model performance. Experimental results show that baseline models using gemini-2.5-flash generally outperform those using gpt-3.5, reflecting differences in LLM capabilities on medical-domain tasks. Notably, when gemini-2.5-flash is combined with the OEMA framework, its performance potential is further unlocked: the exact-match F1 score surpasses that of the traditional CRF method by 4\%–6\%. Perhaps most interestingly, although the gemini backbone yields stronger baseline performance, the performance gains introduced by OEMA are larger when gpt-3.5 is used, suggesting that OEMA may offer greater optimization for relatively weaker LLM backbones.

\begin{table}[htbp]
\centering
\caption{MT Samples and VAERS Relaxed-Match Results.}
\label{tab:main_results_2}
\adjustbox{width=\columnwidth}{
\begin{tabular}{l l c c c c c c}
\toprule
 & & \multicolumn{3}{c}{\textbf{MT Samples}} & \multicolumn{3}{c}{\textbf{VAERS}} \\
\cmidrule(lr){3-5} \cmidrule(lr){6-8}
\textbf{Backbones} & \textbf{Models} & \textbf{P} & \textbf{R} & \textbf{F1} & \textbf{P} & \textbf{R} & \textbf{F1} \\
\midrule
gpt-3.5-turbo & \multirow{2}{*}{OEMA (ours)} & 78.4 & 85.9 & 82.0 & 65.3 & 69.8 & 67.5 \\
gemini-2.5-flash & & 83.1 & 92.2 & 87.4 & 72.3 & 82.8 & 77.2 \\
\midrule
\multirow{2}{*}{supervised learning} & CRF & 66.2 & 88.7 & 75.8 & 60.9 & 76.4 & 67.8 \\
 & BioClinicalBERT & 91.5 & 88.7 & 90.1 & 84.6 & 76.1 & 80.2 \\
\bottomrule
\end{tabular}
}
\end{table}

We now turn to RQ2, examining the performance of OEMA on zero-shot clinical NER tasks compared to traditional supervised learning. BioClinicalBERT, a model specifically fine-tuned on biomedical-domain text with clearly annotated entity boundaries, contrasts with LLMs, which are pretrained on much broader, more diverse general corpora. This fundamental difference in training strategy can lead to varying abilities in boundary detection when handling clinical NER tasks—particularly in highly specialized clinical texts with complex medical terminology\cite{b15_hu2024improving}. As shown in Table.~\ref{tab:main_results_1}, under the exact-match evaluation, OEMA with gemini-2.5-flash as its LLM backbone trails BioClinicalBERT by 10\%–14\% in F1 score, prompting us to further investigate performance under the relaxed-match scenario.

The relaxed-match experimental results in Table.~\ref{tab:main_results_2} reveal even more valuable insights. Under the relaxed-match evaluation criterion, the OEMA framework built on gpt-3.5 already demonstrates performance on par with the traditional CRF method. It is noteworthy that, although BioClinicalBERT still achieves the best results across all datasets, this advantage rests on supervised training with a large volume of manually annotated data. Specifically, BioClinicalBERT’s F1 score is only about 3\% higher than that of the gemini-2.5-flash–based OEMA framework, while the latter achieves a substantial 8\%–10\% improvement over the CRF model. These figures indicate that, in a fully zero-shot setting and without any domain-specific annotations, the OEMA framework can match the performance of the supervised BioClinicalBERT model and significantly outperform the traditional machine learning methods like CRF.

\subsubsection{Ablation Experiment} \label{sec:3.4.2}

\begin{table*}[hbtp]
    \centering
    \caption{MTSamples and VAERS Ablation Test Results, Numbers in \textbf{bold} are the highest results for the corresponding dataset, while numbers \underline{underlined} represent the second-best results.}
    \label{tab:results_ablation}
    \begin{tabularx}{\textwidth}{l l *{12}{>{\centering\arraybackslash}X}}
        \toprule
        & & \multicolumn{6}{c}{MTSamples} & \multicolumn{6}{c}{VAERS} \\
        \cmidrule(lr){3-8} \cmidrule(lr){9-14}
        & Models & \multicolumn{3}{c}{Exact match} & \multicolumn{3}{c}{Relaxed match} & \multicolumn{3}{c}{Exact match} & \multicolumn{3}{c}{Relaxed match} \\
        \cmidrule(lr){3-5} \cmidrule(lr){6-8} \cmidrule(lr){9-11} \cmidrule(lr){12-14}
        Backbones & & P & R & F1 & P & R & F1 & P & R & F1 & P & R & F1 \\
        \midrule
        \multirow{4}{*}{gpt-3.5-turbo} & SILM & 48.0 & 43.1 & 45.4 & \textbf{82.7} & 74.2 & 78.2 & 39.1 & 36.7 & 37.9 & \underline{61.7} & 57.8 & \underline{59.7} \\
        & OEMA (ours) & \underline{49.4} & \textbf{54.1} & \textbf{51.6} & 78.4 & \textbf{85.9} & \textbf{82.0} & \textbf{46.4} & \textbf{49.6} & \textbf{48.0} & \textbf{65.3} & \textbf{69.8} & \textbf{67.5} \\
        & \quad -- No Entity type description & \textbf{50.2} & \underline{52.3} & \underline{51.2} & \underline{78.6} & \underline{82.0} & \underline{80.3} & 41.7 & 41.4 & 41.5 & 59.5 & \underline{59.0} & 59.3 \\
        & \quad -- No Examples & 46.9 & 45.9 & 46.4 & 78.0 & 76.3 & 77.1 & \underline{44.7} & \underline{43.1} & \underline{43.9} & 60.0 & 57.8 & 58.9 \\
        \bottomrule
    \end{tabularx}
\end{table*}

To better understand the individual contributions of the entity-type description prompt and the self-annotated few-shot prompt in the OEMA framework, we address RQ3. Using GPT-3.5 as the LLM backbone, we conduct controlled experiments on two medical datasets (MTSamples and VAERS) and report detailed results in Tables~\ref{tab:results_ablation}. From the results, we can observe that both the entity-type descriptions and the self-annotated few-shot prompts play important roles in OEMA.

\begin{itemize}
\item \textbf{No Entity type description}. Under this setting, OEMA’s performance on both datasets dropped noticeably, yet it still demonstrated strong entity recognition capabilities. Specifically, on the MTSamples dataset, the strict‐match F1 score remained 5.81\% higher than the SILLM baseline; on the VAERS dataset, the exact‐match F1 score held a 3.64\% advantage. However, under the relaxed‐match evaluation, performance on VAERS declined by 0.45\% compared to SILLM. This phenomenon reveals an interesting finding: although the self‐annotated few‐shot prompting strategy effectively preserves precise entity boundary detection (exact-match), in the absence of semantic guidance from entity‐type descriptions it impairs the model’s ability to generalize and recognize entity variants (relaxed-match).
\item  \textbf{No Examples}. This configuration led to a more pronounced performance decline: on the MTSamples dataset, the exact-match F1 score fell by 5.2\% and the relaxed-match score by 5.9\%; on the VAERS dataset, exact-match dropped by 4.1\%, and relaxed-match plummeted by 8.6\%. These results compellingly demonstrate the foundational role of the self-annotated few-shot prompting strategy in supporting overall model performance. In particular, without concrete example guidance, the model struggles to accurately capture the specific expression patterns and contextual features of medical entities, leading to a broad limitation in recognition ability. This finding also confirms the constraints of non-reasoning language models when deprived of exemplar references.

\end{itemize}

\subsubsection{Hyperparameter Tuning} \label{sec:3.4.3}

\begin{figure}[htbp]
    \centering
    \includegraphics[width=.9\linewidth]{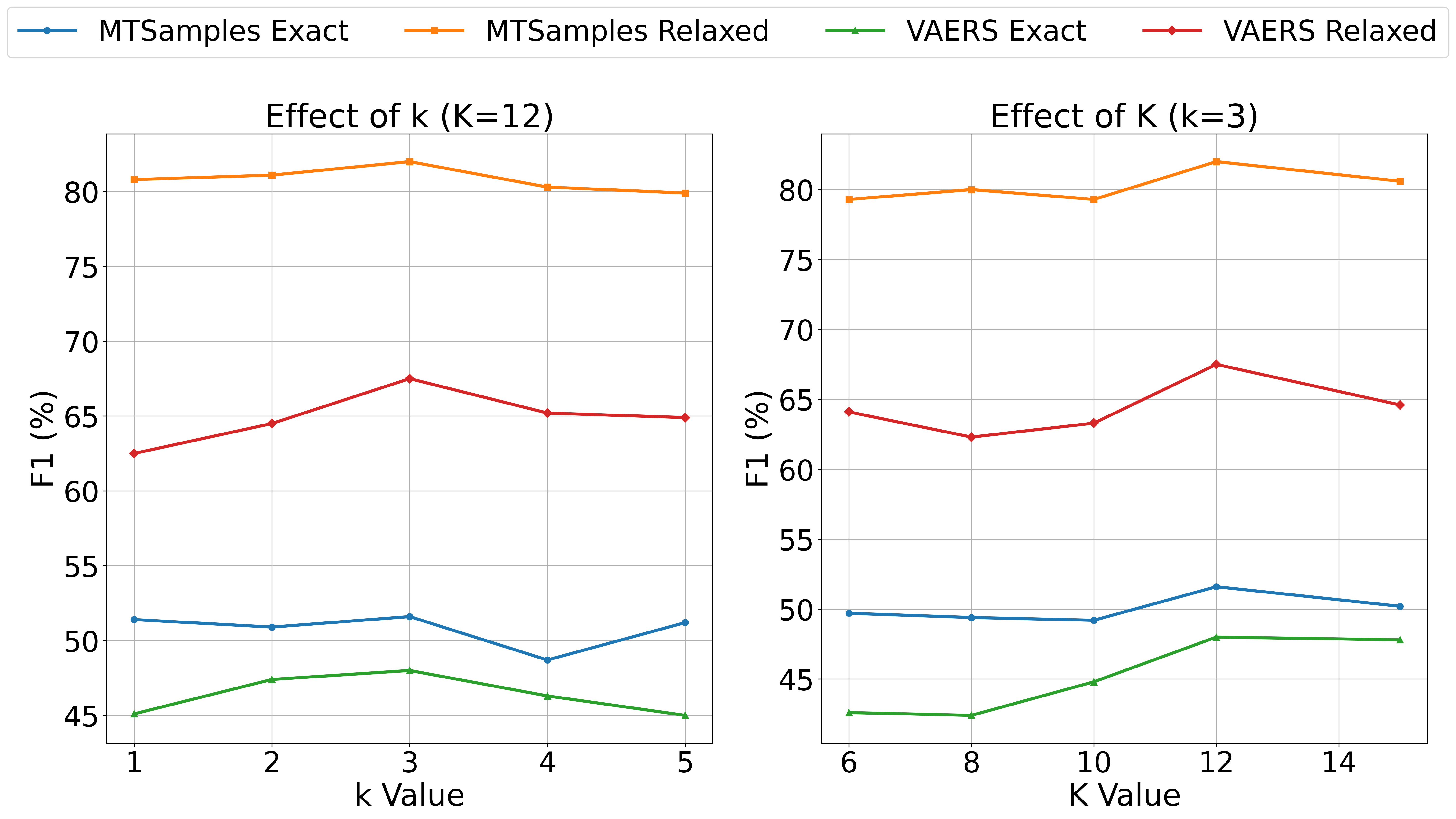}
    \caption{F1 scores (\%) of Diversified KNN under different $k$ (left, with $K{=}12$) and $K$ (right, with $k{=}3$) settings on MTSamples and VAERS.}
    \label{fig:HT}
\end{figure}

\begin{figure*}[htbp]
    \centering
    \includegraphics[width=\linewidth]{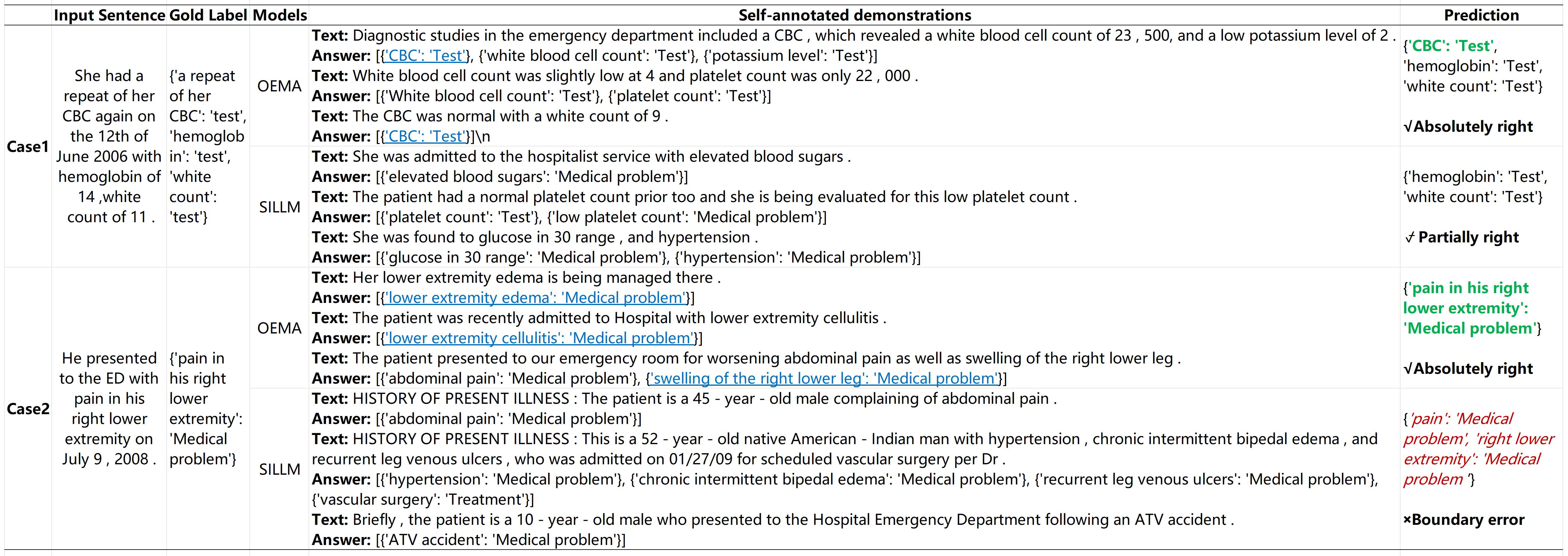}
    \caption{Two specific case analyses. Text marked in bold green indicates entities corrected by OEMA, text marked in italic red indicates incorrect entities, and text marked in underlined blue indicates entities from high-quality self-annotation examples that may help with error correction.}
    \label{fig:CS}
\end{figure*}

We now turn to RQ4, focusing on the tuning process and underlying mechanisms of two key hyperparameters in the OEMA framework’s diversified nearest-neighbor method: $K$ (the candidate‐set size) and $k$ (the number of final examples). As shown in Fig.~\ref{fig:HT}, under the configuration using gpt-3.5 as the LLM backbone, we explored the performance of different parameter combinations under both strict and lenient matching evaluation criteria. The experiments revealed that the OEMA framework’s performance exhibits a clear “bell-shaped” characteristic with respect to these two hyperparameters, indicating a distinct optimal parameter range. When fixing the candidate-set size at $K=12$ and comparing $k\in\{1,2,3,4,5\}$, $k=3$ achieves the best balance under both matching modes; conversely, when fixing the final example count at $k=3$ and varying $K\in\{6,8,10,12,15\}$, $K=12$ demonstrates the most stable performance advantage. This tuning process established the optimal configuration for the aforementioned experiments: a diversified nearest-neighbor candidate-set size of $K=12$ and a final self-labeled example count of $k=3$.

A deeper analysis of these two hyperparameters’ mechanisms shows that $k$ directly determines the number of examples available for the model at inference, requiring a delicate balance. A too-small $k$ (e.g., $k=1$) severely limits generalization, as the few examples cannot cover the diversity of clinical entity expressions;  conversely, a too-large $k$ (e.g., $k=5$) introduces excessive noise from inevitably imperfect self-annotations, degrading prediction quality. Equally crucial is the choice of $K$, which governs the search space for candidate examples and also demands careful trade-offs. If $K$ is too small (e.g., $K=6$), the retrieval space becomes constrained, possibly missing truly high-quality references; in the extreme case when $K=k$, the diversification mechanism collapses into a standard $k$-nearest-neighbor algorithm, losing its diversity guarantees. If $K$ is too large (e.g., $K=15$), although the candidate pool expands, the discriminator faces increased selection pressure, raising computational overhead and potentially lowering the final set’s quality.

\subsubsection{Case Study} \label{sec:3.4.4}

We now turn to RQ5 to delve into the specific mechanisms and implementation processes by which the OEMA framework enhances performance. We selected two representative test target sentences and, based on the gpt-3.5 model architecture, conducted a fine-grained comparison of OEMA against the baseline method SILLM. Fig.~\ref{fig:CS} clearly illustrates how OEMA improves prediction quality through example-guided prompting: by leveraging its innovative self-annotation mechanism to generate high-quality examples, OEMA successfully corrects recognition errors that occurred in SILLM.

A closer inspection shows that OEMA’s performance improvements stem primarily from its novel, ontology-enhanced strategy for selecting token-level examples. Unlike traditional sentence-level filtering, this fine-grained approach elevates matching precision from the sentence level down to individual tokens, ensuring that the selected examples are highly relevant—both in entity type and contextual features—to the current recognition task. Such precise matching markedly enhances the guiding effect of examples on the LLM’s predictions, allowing the model to learn correct recognition patterns from the most pertinent reference cases. By contrast, traditional sentence-level example-selection methods suffer clear limitations: whole-sentence similarity often fails to capture the specific matching needs of entity recognition tasks, and imprecise filtering may introduce irrelevant or even incorrect self-annotated samples, whose noise can disrupt the model’s predictive judgments.

\section{Discussion}

This study presents OEMA, an ontology-enhanced multi-agent framework designed to advance zero-shot clinical named entity recognition (NER). By integrating ontology-guided reasoning with collaborative agent interactions, OEMA addresses two persistent challenges in zero-shot clinical NER: the mismatch between example granularity and task objectives, and the limited integration between prompt design and self-improvement mechanisms. The modular architecture—consisting of a self-annotator, discriminator, and predictor.

Empirical results demonstrate that OEMA outperforms existing zero-shot baselines and achieves performance comparable to supervised models under relaxed evaluation criteria. The ontology-enhanced discriminator effectively filters pseudo-examples at the token level, reducing noise in self-annotated data and improving semantic alignment with clinical concepts. Compared with prior prompt-based approaches that rely on handcrafted examples or fixed templates, OEMA provides an automated and flexible mechanism for adaptive corpus generation and prompt optimization. These features highlight its potential to reduce reliance on manual annotations while maintaining clinical relevance and interpretability. Importantly, the framework exhibits robustness across multiple backbone large language models (LLMs), including GPT-3.5 and Gemini 2.5, demonstrating that its design can generalize well across different model architectures and domains.

Nevertheless, several limitations should be acknowledged. The current evaluation is based on two relatively small datasets, which may not capture the full diversity of clinical language. The choice of smaller datasets was primarily due to cost and resource constraints associated with large-scale data annotation. Further validation on larger, more heterogeneous corpora is necessary to confirm robustness and generalizability. Moreover, the dependence on SNOMED CT constrains the framework’s applicability to domains lacking comprehensive ontologies.

Beyond these aspects, several promising research directions emerge for OEMA. One significant avenue involves integrating continual learning mechanisms that enable agents to iteratively refine their reasoning and decision-making through user feedback, adaptive supervision, or evolving clinical data streams. Such mechanisms would transform OEMA from a static, zero-shot framework into a dynamic, self-improving ecosystem capable of maintaining performance and relevance in complex, real-world clinical environments. Future research should also focus on broadening OEMA’s applicability to open-domain clinical NER tasks, where diverse terminologies and context-dependent meanings challenge current ontology-based methods. Incorporating hybrid symbolic–neural reasoning could help balance interpretability with flexibility, mitigating overreliance on predefined ontologies while allowing the system to generalize more effectively. Moreover, extending OEMA beyond NER to downstream tasks—such as relation extraction, entity linking, and clinical event detection—could further establish it as a unified foundation for ontology-enhanced clinical NLP, capable of supporting richer, context-aware understanding across the entire clinical text analysis pipeline.

In summary, OEMA demonstrates that ontology-guided multi-agent collaboration can enhance zero-shot clinical NER.  While preliminary results are promising, further studies on scalability, real-world deployment, and human-in-the-loop evaluation are warranted to realize its full potential in clinical natural language processing.

\section{Conclusion}
To address the mismatch between example selection and task granularity, as well as the lack of practical integration between prompt design and self-improvement frameworks — both in the context of zero-shot clinical named entity recognition — this study proposes an ontology-enhanced multi-agent framework (OEMA). Experimental results show that under strict matching criteria, the OEMA approach significantly outperforms previous zero-shot learning methods in terms of precision, recall, and F1 score. Moreover, under relaxed matching criteria, OEMA achieves performance close to that of the supervised learning model BioClinicalBERT, while substantially surpassing traditional machine learning methods such as CRF. In future work, we plan to further extend OEMA to support open-domain clinical NER tasks.

\section{Author contributions}
X.Z. conceived and designed the research. X.T. performed the experiments, and analyzed the data. X.T., and X.D. drafted the manuscript. X.T., X.D., and X.Z. were involved in the data curation and analysis. X.Z., and X.D. revised the manuscript. All authors read and approved the final manuscript.

\section{Data availability}
Our code and datasets are available at: \url{https://github.com/XinliTao/OEMA}

\section{Funding}
This work is partially supported by the National Natural Science Foundation of China (Nos. U23B2062, 82374302, 82274352), the National Key Research and Development Program (No. 2023YFC3502604), and the Natural Science Foundation of Beijing (No. L232033).

\section{Conflicts of interest}
None declared.

\bibliographystyle{plain}
\bibliography{reference}

@inproceedings{b1_kollapally2024using,
  title={Using clinical entity recognition for curating an interface terminology to aid fast skimming of EHRs},
  author={Kollapally, Navya Martin and Dehkordi, Mahshad Koohi H and Perl, Yehoshua and Geller, James and Deek, Fadi P and Liu, Hao and Keloth, Vipina K and Elhanan, Gai and Einstein, Andrew J and Zhou, Shuxin},
  booktitle={2024 IEEE International Conference on Bioinformatics and Biomedicine (BIBM)},
  pages={6427--6434},
  year={2024},
  organization={IEEE}
}

@article{b2_landolsi2023information,
  title={Information extraction from electronic medical documents: state of the art and future research directions},
  author={Landolsi, Mohamed Yassine and Hlaoua, Lobna and Ben Romdhane, Lotfi},
  journal={Knowledge and Information Systems},
  volume={65},
  number={2},
  pages={463--516},
  year={2023},
  publisher={Springer}
}

@article{b3_nadkarni2011natural,
  title={Natural language processing: an introduction},
  author={Nadkarni, Prakash M and Ohno-Machado, Lucila and Chapman, Wendy W},
  journal={Journal of the American Medical Informatics Association},
  volume={18},
  number={5},
  pages={544--551},
  year={2011},
  publisher={BMJ Group BMA House, Tavistock Square, London, WC1H 9JR}
}

@article{b4_wang2018clinical,
  title={Clinical information extraction applications: a literature review},
  author={Wang, Yanshan and Wang, Liwei and Rastegar-Mojarad, Majid and Moon, Sungrim and Shen, Feichen and Afzal, Naveed and Liu, Sijia and Zeng, Yuqun and Mehrabi, Saeed and Sohn, Sunghwan and others},
  journal={Journal of biomedical informatics},
  volume={77},
  pages={34--49},
  year={2018},
  publisher={Elsevier}
}

@article{b5_jiang2011study,
  title={A study of machine-learning-based approaches to extract clinical entities and their assertions from discharge summaries},
  author={Jiang, Min and Chen, Yukun and Liu, Mei and Rosenbloom, S Trent and Mani, Subramani and Denny, Joshua C and Xu, Hua},
  journal={Journal of the American Medical Informatics Association},
  volume={18},
  number={5},
  pages={601--606},
  year={2011}
}

@article{b6_huang2015bidirectional,
  title={Bidirectional LSTM-CRF models for sequence tagging},
  author={Huang, Zhiheng and Xu, Wei and Yu, Kai},
  journal={arXiv preprint arXiv:1508.01991
        
        
        },
  year={2015}
}

@inproceedings{b7_alsentzer2019publicly,
  title={Publicly Available Clinical BERT Embeddings},
  author={Alsentzer, Emily and Murphy, John and Boag, William and Weng, Wei-Hung and Jindi, Di and Naumann, Tristan and McDermott, Matthew},
  booktitle={Proceedings of the 2nd Clinical Natural Language Processing Workshop},
  pages={72--78},
  year={2019}
}

@inproceedings{b8_bayat2023survey,
  title={Survey and Experiments on Biomedical Pre-Trained Language Models for Named Entity Recognition},
  author={Bayat, Nooshin and Garc{\'\i}a-Santa, Nuria and Cetina, Kendrick and Mart{\'\i}nez, Ander},
  booktitle={2023 IEEE International Conference on Bioinformatics and Biomedicine (BIBM)},
  pages={4282--4288},
  year={2023},
  organization={IEEE}
}

@article{b9_10.1145/3458754,
  title={Domain-specific language model pretraining for biomedical natural language processing},
  author={Gu, Yu and Tinn, Robert and Cheng, Hao and Lucas, Michael and Usuyama, Naoto and Liu, Xiaodong and Naumann, Tristan and Gao, Jianfeng and Poon, Hoifung},
  journal={ACM Transactions on Computing for Healthcare (HEALTH)},
  volume={3},
  number={1},
  pages={1--23},
  year={2021},
  publisher={ACM New York, NY}
}

@article{b10_fi15060192,
  title={Chatgpt and open-ai models: A preliminary review},
  author={Roumeliotis, Konstantinos I and Tselikas, Nikolaos D},
  journal={Future Internet},
  volume={15},
  number={6},
  pages={192},
  year={2023},
  publisher={MDPI}
}

@article{b11_touvron2023llama,
  title={Llama 2: Open foundation and fine-tuned chat models},
  author={Touvron, Hugo and Martin, Louis and Stone, Kevin and Albert, Peter and Almahairi, Amjad and Babaei, Yasmine and Bashlykov, Nikolay and Batra, Soumya and Bhargava, Prajjwal and Bhosale, Shruti and others},
  journal={arXiv preprint arXiv:2307.09288
        
        
        
        },
  year={2023}
}

@article{b12_chowdhery2023palm,
  title={Palm: Scaling language modeling with pathways},
  author={Chowdhery, Aakanksha and Narang, Sharan and Devlin, Jacob and Bosma, Maarten and Mishra, Gaurav and Roberts, Adam and Barham, Paul and Chung, Hyung Won and Sutton, Charles and Gehrmann, Sebastian and others},
  journal={Journal of Machine Learning Research},
  volume={24},
  number={240},
  pages={1--113},
  year={2023}
}

@inproceedings{b13_xie-etal-2023-empirical,
  title={Empirical Study of Zero-Shot NER with ChatGPT},
  author={Xie, Tingyu and Li, Qi and Zhang, Jian and Zhang, Yan and Liu, Zuozhu and Wang, Hongwei},
  booktitle={Proceedings of the 2023 Conference on Empirical Methods in Natural Language Processing},
  pages={7935--7956},
  year={2023}
}

@inproceedings{b14_zhao2024effective,
  title={Effective In-Context Learning for Named Entity Recognition},
  author={Zhao, Jin and Guo, Qian and Liang, Jiaqing and Li, Zhixu and Xiao, Yanghua},
  booktitle={2024 IEEE International Conference on Bioinformatics and Biomedicine (BIBM)},
  pages={1376--1382},
  year={2024},
  organization={IEEE}
}

@article{b15_hu2024improving,
  title={Improving large language models for clinical named entity recognition via prompt engineering},
  author={Hu, Yan and Chen, Qingyu and Du, Jingcheng and Peng, Xueqing and Keloth, Vipina Kuttichi and Zuo, Xu and Zhou, Yujia and Li, Zehan and Jiang, Xiaoqian and Lu, Zhiyong and others},
  journal={Journal of the American Medical Informatics Association},
  volume={31},
  number={9},
  pages={1812--1820},
  year={2024},
  publisher={Oxford Academic}
}

@article{b16_wei2023chatie,
  title={Chatie: Zero-shot information extraction via chatting with ChatGPT},
  author={Wei, Xiang and Cui, Xingyu and Cheng, Ning and Wang, Xiaobin and Zhang, Xin and Huang, Shen and Xie, Pengjun and Xu, Jinan and Chen, Yufeng and Zhang, Meishan and others},
  journal={arXiv preprint arXiv:2302.10205
        
        
        
        
        
        
        
        
        
        
        
        },
  year={2023}
}

@article{b17_wang2022self,
  title={Self-consistency improves chain of thought reasoning in language models},
  author={Wang, Xuezhi and Wei, Jason and Schuurmans, Dale and Le, Quoc and Chi, Ed and Narang, Sharan and Chowdhery, Aakanksha and Zhou, Denny},
  journal={arXiv preprint arXiv:2203.11171
        
        
        
        
        
        
        
        
        
        
        
        },
  year={2022}
}

@inproceedings{b18_xie-etal-2024-self,
  title={Self-Improving for Zero-Shot Named Entity Recognition with Large Language Models},
  author={Xie, Tingyu and Li, Qi and Zhang, Yan and Liu, Zuozhu and Wang, Hongwei},
  booktitle={Proceedings of the 2024 Conference of the North American Chapter of the Association for Computational Linguistics: Human Language Technologies (Volume 2: Short Papers)},
  pages={583--593},
  year={2024}
}

@article{b19_wang2024reversener,
  title={ReverseNER: A Self-Generated Example-Driven Framework for Zero-Shot Named Entity Recognition with Large Language Models},
  author={Wang, Anbang and Mei, Difei and Zhang, Zhichao and Bai, Xiuxiu and Yao, Ran and Fang, Zewen and Hu, Min and Cao, Zhirui and Sun, Haitao and Guo, Yifeng and others},
  journal={arXiv preprint arXiv:2411.00533
        
        
        
        
        
        
        
        
        
        
        
        
        
        },
  year={2024}
}

@inproceedings{b21_wang-etal-2025-gpt,
  title={GPT-NER: Named Entity Recognition via Large Language Models},
  author={Wang, Shuhe and Sun, Xiaofei and Li, Xiaoya and Ouyang, Rongbin and Wu, Fei and Zhang, Tianwei and Li, Jiwei and Wang, Guoyin and Guo, Chen},
  booktitle={Findings of the Association for Computational Linguistics: NAACL 2025},
  pages={4257--4275},
  year={2025}
}

@inproceedings{b22_zhang-etal-2025-survey,
  title={A Survey of Generative Information Extraction},
  author={Zhang, Zikang and You, Wangjie and Wu, Tianci and Wang, Xinrui and Li, Juntao and Zhang, Min},
  booktitle={Proceedings of the 31st International Conference on Computational Linguistics},
  pages={4840--4870},
  year={2025}
}

@article{b23_uzuner20112010,
  title={2010 i2b2/VA challenge on concepts, assertions, and relations in clinical text},
  author={Uzuner, {\"O}zlem and South, Brett R and Shen, Shuying and DuVall, Scott L},
  journal={Journal of the American Medical Informatics Association},
  volume={18},
  number={5},
  pages={552--556},
  year={2011},
  publisher={BMJ Group BMA House, Tavistock Square, London, WC1H 9JR}
}

@article{b24_du2021extracting,
  title={Extracting postmarketing adverse events from safety reports in the vaccine adverse event reporting system (VAERS) using deep learning},
  author={Du, Jingcheng and Xiang, Yang and Sankaranarayanapillai, Madhuri and Zhang, Meng and Wang, Jingqi and Si, Yuqi and Pham, Huy Anh and Xu, Hua and Chen, Yong and Tao, Cui},
  journal={Journal of the American Medical Informatics Association},
  volume={28},
  number={7},
  pages={1393--1400},
  year={2021},
  publisher={Oxford University Press}
}

@article{savova2010mayo,
  title={Mayo clinical Text Analysis and Knowledge Extraction System (cTAKES): architecture, component evaluation and applications},
  author={Savova, Guergana K and Masanz, James J and Ogren, Philip V and Zheng, Jiaping and Sohn, Sunghwan and Kipper-Schuler, Karin C and Chute, Christopher G},
  journal={Journal of the American Medical Informatics Association},
  volume={17},
  number={5},
  pages={507--513},
  year={2010},
  publisher={BMJ Group BMA House, Tavistock Square, London, WC1H 9JR}
}

@inproceedings{devlin2019bert,
  title={Bert: Pre-training of deep bidirectional transformers for language understanding},
  author={Devlin, Jacob and Chang, Ming-Wei and Lee, Kenton and Toutanova, Kristina},
  booktitle={Proceedings of the 2019 conference of the North American chapter of the association for computational linguistics: human language technologies, volume 1 (long and short papers)},
  pages={4171--4186},
  year={2019}
}

@article{lee2020biobert,
  title={BioBERT: a pre-trained biomedical language representation model for biomedical text mining},
  author={Lee, Jinhyuk and Yoon, Wonjin and Kim, Sungdong and Kim, Donghyeon and Kim, Sunkyu and So, Chan Ho and Kang, Jaewoo},
  journal={Bioinformatics},
  volume={36},
  number={4},
  pages={1234--1240},
  year={2020},
  publisher={Oxford University Press}
}

@article{huang2019clinicalbert,
  title={Clinicalbert: Modeling clinical notes and predicting hospital readmission},
  author={Huang, Kexin and Altosaar, Jaan and Ranganath, Rajesh},
  journal={arXiv preprint arXiv:1904.05342
        
        
        
        
        
        
        
        
        
        
        
        
        
        },
  year={2019}
}
\onecolumn
\begin{appendices}
\setcounter{table}{0}
\renewcommand{\thetable}{A\arabic{table}}
\section*{Supplementary Material}\label{sec11}

\begin{table}[htbp]
    \centering
    \caption{
    An example of the zero-shot in-context learning (ICL) prompt used in the self-annotation stage of the OEMA framework. 
    The prompt guides the LLM to identify named entities in raw clinical text based on a predefined entity type and output them in a JSON format. 
    The resulting self-annotated corpus is then filtered via ontology-enhanced token-level example selection to obtain high-quality few-shot exemplars for the final NER inference.}
    \label{tab:prompt_zero_shot}
    \begin{tabular}{p{0.95\linewidth}}
    \toprule \textbf{Prompt designed for self-annotation} \end{tabular}
    \begin{tabular}{p{0.95\linewidth}} 
        \toprule
        You are an expert in medical named entity recognition. You're very good at extracting information. Given entity label set: [‘Medical problem’, ‘Treatment’, ‘Test’] \\\\
        Please recognize the named entities in the given text. Based on the given entity label set, provide answer in the following JSON format: [\{‘Entity Name’: ‘Entity Label’\}]. If there is no entity in the text, return the following empty list: []. Only return answer, not explanations. \\\\
        Text: “The patient presented to our emergency room for worsening abdominal pain as well as swelling of the right lower leg.” \\
        Answer:\\
        \bottomrule
    \end{tabular}
\end{table}

\begin{table}[htbp]
    \centering
    \caption{
    An example of the ICL prompt used for extracting clinical ontologies based on the SNOMED CT top-level hierarchy. 
    The prompt instructs the LLM to identify medical concepts appearing in the input text according to the 18 top-level SNOMED CT categories, and to output results in a JSON format. 
    These ontology-level annotations are later used by the discriminator to evaluate token-level similarity and assist in selecting the most relevant examples for downstream NER.}
    \label{tab:prompt_ontology_extraction}
    \begin{tabular}{p{0.95\linewidth}}
    \toprule \textbf{Prompt designed for extracting the clinical ontologies} \end{tabular}
    \begin{tabular}{p{0.95\linewidth}} 
        \toprule
        Please refer to the 18 top-level categories defined in the Concept Hierarchy of SNOMED CT, and explicitly extract the clinical medical ontologies mentioned in the text in the order of their appearance.\\\\
        Provide answer in the format: \{“(top-level category, ontology)”: “original text fragment”, ...\}. Only a dictionary string should be returned, without any Markdown formatting, code blocks, or additional content.\\\\
        Text: “She started off with a little pimple on the buttock.”\\
        Answer: \{“(Clinical finding, Pustule)”: “pimple”, “(Body structure, Buttock)”: “buttock”\}\\
        \dots\dots\\
        (Selected examples with medicine ontology)\\
        \dots\dots\\
        Text: The patient presented to our emergency room for worsening abdominal pain as well as swelling of the right lower leg.\\
        Answer:\\
        \bottomrule
    \end{tabular}
\end{table}

\begin{table}[htbp]
    \centering
    \caption{
    An example of the ICL prompt used in the example-scoring stage of the OEMA framework. 
    The prompt guides the LLM to assign helpfulness scores to candidate self-annotated examples by evaluating the relevance of their clinical ontology–text fragment pairs to those in the target sentence. 
    This scoring process enables the discriminator to select the most suitable few-shot exemplars for constructing the final inference prompt.}
    \label{tab:prompt_examples_scoring}
    \begin{tabular}{p{0.95\linewidth}}
    \toprule \textbf{Prompt designed for scoring examples} \end{tabular}
    \begin{tabular}{p{0.95\linewidth}} 
        \toprule
        \#\#\# Example Scoring for Entity Recognition Tasks\\
        Given entity label set: [‘Medical problem’, ‘Treatment’, ‘Test’] and target sentence: \{\\
        ‘sentence’: ‘She started off with a little pimple on the buttock.’, \\ 
        ‘ontology’: ‘\{“(Clinical finding, Pustule)”: “pimple”, “(Body structure, Buttock)”: “buttock”\}’\\
        \}\\\\
        \#\#\# Scoring Guidelines\\
        Based on the target sentence has learned SNOMED CT medical ontology and may involve entity type, please predict the helpfulness scores and give reasons of each sentence, which indicates the degree to which providing the current sentence can aid in extracting named entities from the target\_sentence. The score ranges from 1 to 5, with 1 being the least helpful and 5 being the most helpful.\\\\
        Provide answer in the following JSON format: [\{“idx”: “sentence identifier”, “score”: “be strict and reflect the differences in scores, not all 1 or all 5”, “reason”: “combined with the characteristics of the target sentence”\}, ...]\\
        Make sure that the output is a complete string, do not use newline characters, Markdown format, \texttt{```json}, or any additional instructions, and only return formatted string results.\\
        \dots\dots\\
        (Selected examples with medicine ontology)\\
        \dots\dots\\
        \bottomrule
    \end{tabular}
\end{table}

\begin{table}[htbp]
    \centering
    \caption{
    An example of the ICL prompt used in the final prediction stage of the OEMA framework. 
    The prompt integrates entity-type descriptions with the selected high-quality self-annotated examples to provide structured, type-aware guidance for the LLM. 
    Based on this combined prompting, the LLM performs the final clinical named entity recognition on the target sentence and outputs the results in a JSON format.}
    \label{tab:prompt_predict}
    \begin{tabular}{p{0.95\linewidth}}
    \toprule \textbf{Prompt designed for the final prediction} \end{tabular}
    \begin{tabular}{p{0.95\linewidth}} 
        \toprule
        You are an expert in medical named entity recognition. You're very good at extracting information.\\
        \dots\dots\\
        (Entity type description)\\
        \dots\dots\\
        Given entity label set: [‘Medical problem’, ‘Treatment’, ‘Test’]\\
        Please recognize the named entities in the given text. Based on the given entity label set, provide answer in the following JSON format: [\{‘Entity Name’: ‘Entity Label’\}]. If there is no entity in the text, return the following empty list: []. Only return answer, not explanations. \\\\
        Text: “She would usually have pustular type of lesion that would eventually break and would be quite painful.”\\
        Answer: [\{“pustular type of lesion”: “Medical problem”\}]\\
        \dots\dots\\
        (Selected examples with self-annotated label)\\
        \dots\dots\\
        Text: “She started off with a little pimple on the buttock.”\\
        Answer:\\
        \bottomrule
    \end{tabular}
\end{table}

\end{appendices}
\end{document}